\documentclass[10pt,twocolumn,letterpaper]{article}

\usepackage{cvpr}              

\usepackage[accsupp]{axessibility} 

\usepackage{bm}
\newcommand{\cparagraph}[1]{{\vspace{+1mm}\noindent\textbf{#1}}}

\usepackage[dvipsnames]{xcolor}
\definecolor{cvprblue}{rgb}{0.21,0.49,0.74}
\usepackage[pagebackref,breaklinks,colorlinks,citecolor=cvprblue]{hyperref}
\usepackage{float}
\usepackage{marvosym}
\usepackage{amsmath}

\title{From Parts to Whole: A Unified Reference Framework for Controllable Human Image Generation}

\author{
    {
        Zehuan Huang\footnotemark[1] \quad Hongxing Fan\footnotemark[1] \quad Lipeng Wang\footnotemark[1] \quad Lu Sheng\footnotemark[2]
    }\\
    {Beihang University} \\
    {\tt\small \{huangzehuan, fanhongxing, wanglipeng, lsheng\}@buaa.edu.cn}\\
    {\small\url{https://huanngzh.github.io/Parts2Whole/}}
}

\begin{document}

\twocolumn[\maketitle\vspace{-2.5em}\begin{center}
\centering
\includegraphics[width=\textwidth]{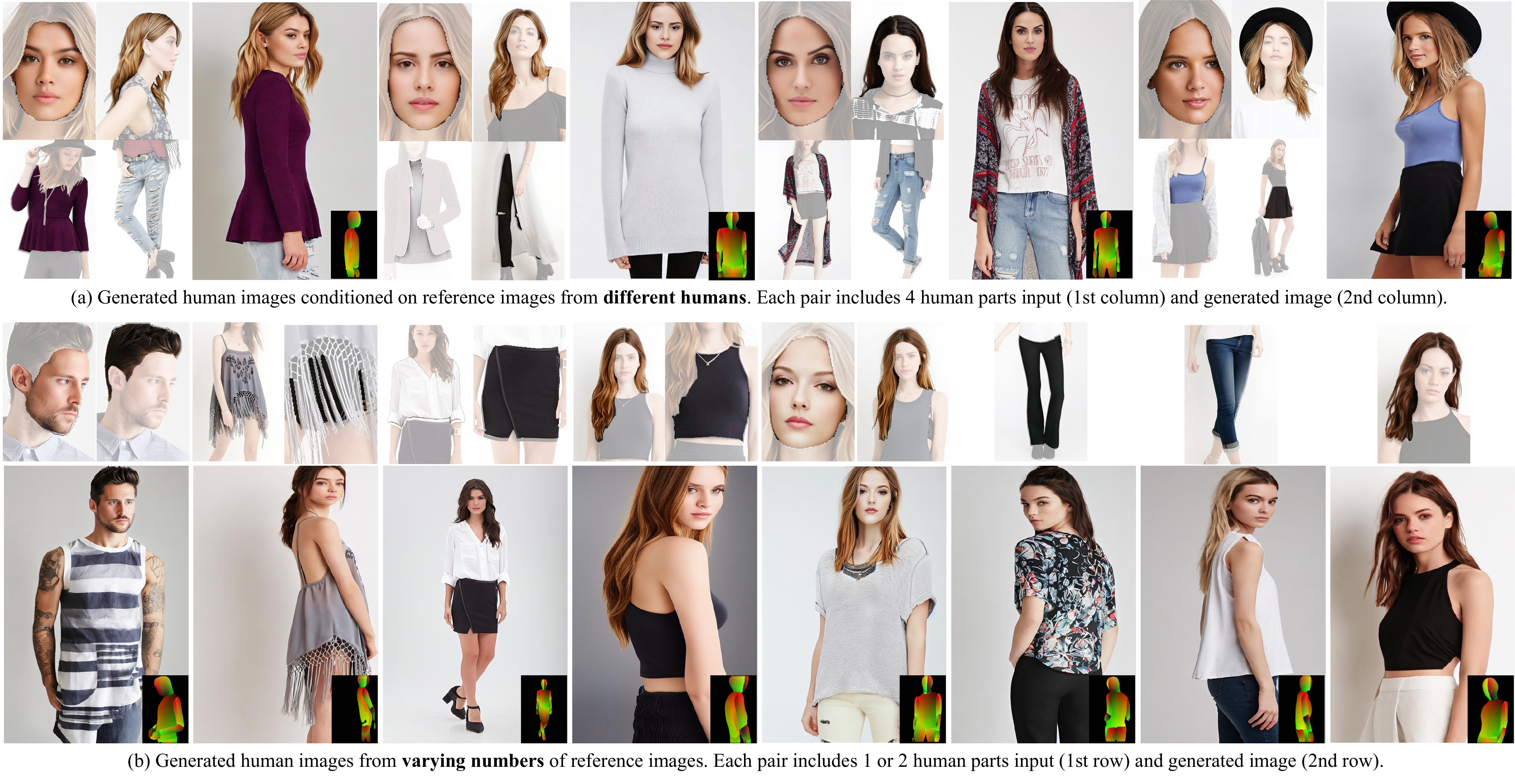}
\captionof{figure}{We propose \textbf{Parts2Whole}, which can generate realistic and high-quality human figures in various postures from referential human part images of any quantity and different origins. Our method maintains the high alignment with the corresponding conditional semantic regions, while ensuring diversity and harmony among the whole body.}
\label{fig:teaser}
\end{center}\bigbreak]

\renewcommand{\thefootnote}{\fnsymbol{footnote}}
\footnotetext[1]{Equal Contribution.}
\footnotetext[2]{Corresponding author.}

\begin{abstract}
Recent advancements in controllable human image generation have led to zero-shot generation using structural signals (e.g., pose, depth) or facial appearance. Yet, generating human images conditioned on multiple parts of human appearance remains challenging. Addressing this, we introduce Parts2Whole, a novel framework designed for generating customized portraits from multiple reference images, including pose images and various aspects of human appearance. To achieve this, we first develop a semantic-aware appearance encoder to retain details of different human parts, which processes each image based on its textual label to a series of multi-scale feature maps rather than one image token, preserving the image dimension. Second, our framework supports multi-image conditioned generation through a shared self-attention mechanism that operates across reference and target features during the diffusion process. We enhance the vanilla attention mechanism by incorporating mask information from the reference human images, allowing for the precise selection of any part. Extensive experiments demonstrate the superiority of our approach over existing alternatives, offering advanced capabilities for multi-part controllable human image customization.
\end{abstract}
\section{Introduction}

Controllable human image generation aims to synthesize human images that align with specific textual descriptions, structural signals or more precise appearance conditions. It emerges as a significant technology within the realm of digital content creation, providing users with a portrait customization solution. However, due to the complexity of the control conditions, this task presents significant challenges, especially when it comes to multi-type condition input and control of various aspects of human appearance.

As diffusion models~\cite{ho2020ddpm,rombach2022ldm,ramesh2022dalle2,saharia2022imagen,nichol2022glide,dhariwal2021diffusionbeatgans} have brought great success in image generation, the task of controllable human image generation has experienced rapid development. Several works~\cite{jiang2022text2human} utilize languages as condition, generating human images by providing attributes about the textures of clothes. Due to the rough control of texts, it struggles to accurately guide the generation of human appearance. Another group of works~\cite{zhang2023controlnet, mou2023t2iadapter, zhao2024unicontrolnet, liu2023hyperhuman} focuses on introducing structural signals to control human posture. Although these methods have achieved impressive results, they do not consider appearance as a condition, which is crucial for portrait customization.

Recently, several works~\cite{ruiz2023dreambooth, hu2021lora, gal2022textualinversion, kumari2023customdiffusion, liu2023cones, shi2023instantbooth, ye2023ipadapter, chen2023anydoor, li2023photomaker, wang2024instantid, zhang2024ssrencoder} have emerged that use appearance conditions to guide human image generation. They learn human representation from reference images and generate images aligning with the specific face identity. One prominent approach involves test-time fine-tuning~\cite{ruiz2023dreambooth, hu2021lora, gal2022textualinversion, kumari2023customdiffusion, liu2023cones}. It requires substantial computational resources to learn each new individual, which costs about half an hour to achieve satisfactory results. Another approach~\cite{shi2023instantbooth, ye2023ipadapter, chen2023anydoor, li2023photomaker, wang2024instantid, zhang2024ssrencoder} investigates the zero-shot setting to bypass the fine-tuning cost. It encodes the reference image into one or several tokens and injects them into the generation process along with text tokens. These zero-shot methods make human image customization practical with faster speed. However, due to the loss of spatial representations when encoding the reference images into one or a few tokens, they struggle to preserve appearance details. And they lack the design to obtain specified information from the images, but instead utilize all the information, resulting in ambiguous subject representation.

In this paper, we target generating human images from multi-part images of human appearance, along with specific pose maps or optionally text descriptions. The above-mentioned generation methods conditioned on structural signals~\cite{zhang2023controlnet, mou2023t2iadapter} or face identity~\cite{ruiz2023dreambooth, hu2021lora, gal2022textualinversion, ye2023ipadapter, wang2024instantid, zhang2024ssrencoder} have their limitations on this task (results shown in \cref{fig:comparison} and \cref{fig:ablation}). It is attributed to the spatial misalignment of the input multi-body parts with the target image, and the lack of specific design in existing methods to \textbf{address the variation in spatial positions during feature injection}. Methods like IP-Adapter~\cite{ye2023ipadapter} and SSR-Encoder~\cite{zhang2024ssrencoder} incorporate features into the denoising U-Net through cross-attention mechanisms. They encode reference images into other modal features (e.g., semantic features) and utilize the cross-attention keys and values from them rather than from \textbf{image dimensional} feature maps. As a result, the spatial relationship in the original image dimensions between the reference images and the target image is lost, resulting in a mixture of attributes from different subjects. Although methods like ControlNet~\cite{zhang2023controlnet} encode the reference images into image-dimensional features, they add the features to the feature maps in the U-Net decoder. It is suitable for tasks where the condition maps and the target map have the same structure, such as guiding generation using line drawings. However, in the case of the spatial misalignment of the conditional images with the target image, it is difficult to model the correlation of spatial information by \textbf{directly adding or concat features} on the channel dimension.

To address the above issues, we present Parts2Whole, a unified reference framework for portrait customization from multiple reference images, including various parts of human appearance (e.g., hair, face, clothes, shoes, etc.) and pose maps. Inspired by the effective reference mechanism used in image-to-video tasks~\cite{hu2023animateanyone, xu2023magicanimate}, we develop a semantic-aware appearance encoder based on the Reference U-Net architecture. It encodes each image with its textual label into \textbf{a series of multi-scale feature maps in image dimension}, preserving appearance details and spatial information of multiple reference images. The additional semantic condition represents a category instruction, which helps retain richer shapes and detailed attributes of each aspect. Furthermore, to preserve the positional relationship when injecting reference features into the image generation process, we employ a \textbf{shared self-attention} operation across reference and target features during the diffusion process. We also build a tiny convolution network to extract the pose features and inject them into the generation. To precisely select the specified part from each reference image, we enhance the vanilla self-attention mechanism by \textbf{incorporating masks} of the subjects in the reference images.

Equipped with these techniques, Parts2Whole demonstrates superior quality and controllability for human image generation. Our contributions are summarized as follows:

\begin{itemize}
\item We construct a novel framework, Parts2Whole, which supports the controllable generation of human images conditioned on texts, pose signals, and multiple aspects of human appearance.
\item We propose an advanced multi-reference mechanism consisting of a semantic-aware image encoder and the shared attention operation, which retains details of the specific key elements and achieves precise subject selection with the help of our proposed mask-guided approach.
\item Experiments show that our Parts2Whole generates high-quality human images from multiple conditions and maintains high consistency with the given conditions.
\end{itemize}
\section{Related Work}

\cparagraph{Text-to-Image Generation.}
In recent years, text-to-image generation has made remarkable progress, particularly with the development of diffusion models~\cite{ramesh2022dalle2, nichol2022glide, rombach2022ldm, saharia2022imagen, dhariwal2021diffusionbeatgans, ho2020ddpm, podell2023sdxl, huang2023epidiff} and auto-regressive
models~\cite{chang2023muse, yu2022scalingauto, tian2024var}, which have propelled text-to-image generation to large-scale commercialization. Since DALLE2~\cite{ramesh2022dalle2}, Stable Diffusion~\cite{rombach2022ldm} and Imagen~\cite{saharia2022imagen} employ diffusion models as generative models and train the models on large datasets, text-to-image synthesis ability has been significantly enhanced. More recently, Stable Diffusion XL~\cite{podell2023sdxl}, a two-stage cascade diffusion model, has greatly improved the generation of high-frequency details and overall image color, taking aesthetic appeal to a higher level. However, these existing methods are limited to generating images solely from text prompts, and they do not meet the demand for producing customized images with the preservation of appearance.

\cparagraph{Controllable Image Generation.}
Given the robust generative capabilities of image diffusion models, a series of research~\cite{zhang2023controlnet, mou2023t2iadapter, qin2023unicontrol, ruiz2023dreambooth, hu2021lora, ye2023ipadapter, chen2023anydoor, zhang2024ssrencoder} attempts to explore the controllability of image generation, enabling image synthesis guided by multi-modal conditions. Some work~\cite{zhang2023controlnet, mou2023t2iadapter, jiang2023scedit, qin2023unicontrol, zhao2024unicontrolnet, hu2023cocktail} focuses on introducing structural signals such as edges, depth maps, and segmentation maps, to control the spatial structure of generated images. Another group of work~\cite{ruiz2023dreambooth, hu2021lora, gal2022textualinversion, ye2023ipadapter, chen2023anydoor} uses appearance conditions to guide image generation, aiming to generate images aligning with specific concepts like identity and style, known as subject-driven image generation. The methods generally fall into two categories: those requiring test-time fine-tuning and those that do not. Test-time fine-tuning methods~\cite{ruiz2023dreambooth, hu2021lora, gal2022textualinversion, kumari2023customdiffusion, liu2023cones} often optimizes additional text embedding, parameter residuals or direct fine-tune the whole model to fit the specified subject. Although these methods have achieved impressive results, they cost about half an hour to achieve satisfactory results. Fine-tuning-free methods~\cite{shi2023instantbooth, ye2023ipadapter, chen2023anydoor, zhang2024ssrencoder, ma2023subjectdiffusion, gal2023encoderdiff, wei2023elite} typically train an additional encoding network to encode the reference image into embeddings or image prompts. However, due to the loss of spatial representations when encoding the reference images into one or a few tokens, they struggle to preserve appearance details.

\cparagraph{Controllable Human Image Generation.}
In this paper, we mainly focus on controllable human image generation and aim to synthesize human images aligning with specific text prompts, pose signals, and various parts of human appearance. Text2Human~\cite{jiang2022text2human} generates full-body human images using detailed descriptions about the textures of clothes, but is limited by the coarse-grained textual condition. Test-time fine-tuning methods~\cite{ruiz2023dreambooth,hu2021lora,kumari2023customdiffusion} produce satisfactory results, but when it comes to customizing portraits using multiple parts of human appearance, they take much more time to fit each aspect. Recently, methods like IP-Adapter-FaceID~\cite{ye2023ipadapter}, FastComposer~\cite{xiao2023fastcomposer}, PhotoMaker~\cite{li2023photomaker}, and InstantID~\cite{wang2024instantid} show promising results on zero-shot human image personalization. They encode the reference face to one or several tokens as conditions to generate customized images. With the addition of adaptable structural control networks~\cite{zhang2023controlnet, mou2023t2iadapter}, these methods can generate portraits aligned with specified poses and human identities. However, they usually fail to maintain the details of human identities and utilize all the information from a single image, resulting in ambiguous subject representation. These make it difficult to apply these schemes to precisely generation conditioned on multiple parts of the human appearance. In contrast, our Parts2Whole is both generalizable and efficient, and precisely retains details in multiple parts of human appearance.
\section{Method}

\begin{figure*}
    \centering
    \includegraphics[width=\textwidth]{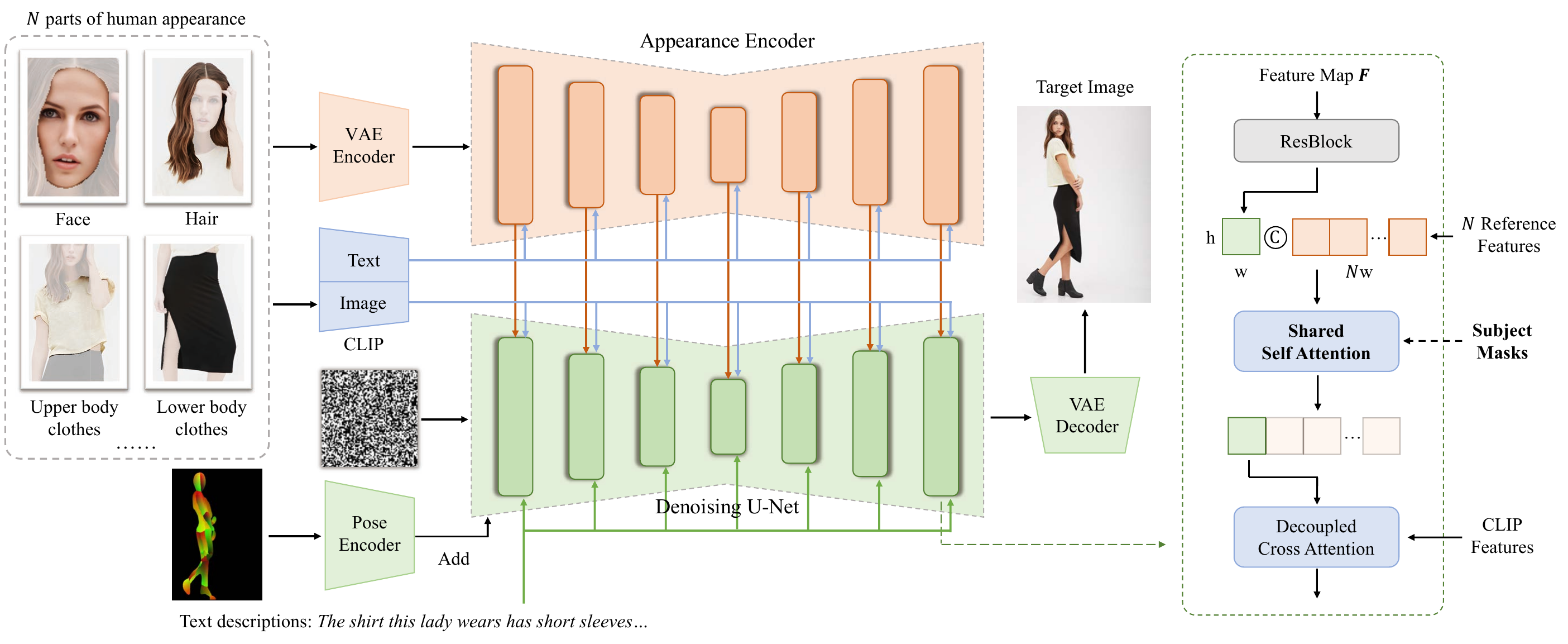}
    \caption{Overview of Parts2Whole. Based on the text-to-image diffusion model, our method designs an appearance encoder for encoding various parts of human appearance into multi-scale feature maps. We build this encoder by copying the network structure and pre-trained weights from denoising U-Net. Features obtained from reference images with their textual labels are injected into the generation process by shared attention mechanism layer by layer. To precisely select the specified parts from reference images, we enhance the vanilla self-attention mechanism by incorporating subject masks in the reference images. An illustration of one block in U-Net is shown on the right part.}
    \label{fig:overview}
\end{figure*}

We target controllable human image generation guided by multiple reference images. Given $N$ images that capture distinct parts of human appearance $\bm{x}^{1:N}$ and a pose map $\bm{p}$, and optionally text inputs, our objective is to synthesize a human image $\hat{\bm{x}}$ aligning with the specified appearances and posture. To achieve this goal, we propose Parts2Whole, a specialized framework designed to interpolate various reference images and generate high-quality portraits.

In general, Parts2Whole is built on text-to-image diffusion models~\cite{rombach2022ldm}. In the following sections, we start with an overview of T2I diffusion models, and in particular, the self-attention mechanism in Sec.~\ref{subsec:preliminaries}. We continue by presenting our unified reference framework in Sec.~\ref{subsec:unifiedref}, which consists of a semantic-aware appearance encoder, a shared self-attention that queries referential features within the self–attention layers, and the enhanced mask-guided subject selection. These methods enable Parts2Whole to accurately obtain the specific subject information from multiple reference images while preserving appearance details.

\subsection{Preliminaries}
\label{subsec:preliminaries}

\cparagraph{Text-to-Image Diffusion Models.}
Diffusion models~\cite{ho2020ddpm} exhibit promising capabilities in image generation. In this study, we select the widely adopted Stable Diffusion~\cite{rombach2022ldm} as our foundational model, which is also known as Latent Diffusion Models (LDM). The model operates the denoising process in the latent space of an autoencoder~\cite{kingma2022ae}, namely $\mathcal{E}(\cdot)$ and $\mathcal{D}(\cdot)$. During the training phase, an input image $\bm{x}_{0}$ is initially mapped to the latent space using a frozen encoder, yielding $\bm{z}_{0}=\mathcal{E}(\bm{x}_{0})$, then perturbed by a pre-defined Markov process:
\begin{equation}
    q(\bm{z}_{t}|\bm{z}_{t-1})=\mathcal{N}(\bm{z}_{t}; \sqrt{1-\beta_{t}}\bm{z}_{t-1}, \beta_{t}\bm{I})
    \label{eq:img_diff}
\end{equation}
For $t=1,\cdots, T$, where $T$ represents the number of steps in the forward diffusion process. The sequence of hyperparameters $\beta_{t}$ determines the noise strength at each step. The denoising UNet $\epsilon_{\theta}$ is trained to approximate the reverse process $q(\bm{z}_{t-1}|\bm{z}_{t})$. The training objective is expressed as:
\begin{equation}
    \mathcal{L}=\mathbb{E}_{\mathcal{E}(\bm{x}_{0}),\bm{\epsilon}\sim\mathcal{N}(\bm{0}, \bm{I}), c, t}[\lVert \bm{\epsilon}-\epsilon_{\theta}(\bm{z}_{t},\bm{c},t) \rVert_{2}^{2}]
    \label{eq:img_diff_loss}
\end{equation}
Here, $\bm{c}$ denotes the conditioning texts. At the inference stage, Stable Diffusion effectively reconstructs an image from Gaussian noise step by step, predicting the noise added at each stage. The denoised results are then fed into a latent decoder $\mathcal{D}(\cdot)$ to regenerate colored images from the latent representations, denoted as $\hat{\bm{x}}_{0} = \mathcal{D}(\hat{\bm{z}}_{0})$.

\cparagraph{Self-Attention in T2I Models.}
Stable Diffusion~\cite{rombach2022ldm} employs a U-Net architecture~\cite{ronneberger2015unet} that consists of convolution layers and transformer attention blocks~\cite{vaswani2017attention}. In these attention mechanisms, self-attention layers are used to aggregate the spatial features of the image itself and cross-attention layers are designed to query information from text embedding. The main difference is that the cross-attention layer uses text features as keys and values, while in self-attention layers, image features with spatial dimensions serve as query, key, and value by themselves, preserving more freedom to represent spatially varying visual elements. The self-attention layer takes a feature map $\bm{F}$ of the image as input and computes the attention of the feature in location $s$ with the entire feature map: 
\begin{equation}
    \tilde{\bm{F}}_{s} = \text{SoftMax}\left(
    \frac{
    Q(\bm{F}_s)\cdot K(\bm{F})^{\top}
    }{
    \sqrt{d}
    }\right)\cdot V(\bm{F})
    \label{eq:diff_self_attn}
\end{equation}
where $Q,K,V$ are linear projection layers, $\bm{F}\in \mathbb{R}^{(hw)\times d}$ is a flattened feature map obtained from the denoiser $\epsilon_{\theta}$, where $d$ is the feature dimension, and $h,w$ are intermediate spatial dimensions. $\bm{F}_s, \tilde{\bm{F}}_s$ is the input and output feature for location $s$ respectively.

Several works extend the self-attention layer to inject the reference image features~\cite{hu2023animateanyone,xu2023magicanimate}, or generate style-aligned or subject-consistent images~\cite{tewel2024consistory, hertz2023stylealigned, jeong2024visualstyleprompt}, and demonstrate the effectiveness of this mechanism. Inspired by them, we extend the keys and values of the self-attention layer to multiple reference images and preserve the details of the referential appearance successfully.

\subsection{Unified Reference Framework}
\label{subsec:unifiedref}

As demonstrated in Fig.~\ref{fig:overview}, our Parts2Whole consists of two branches:  the reference branch used to encode multiple parts of human appearance, and the denoising branch, to gradually denoise the randomly sampled noise to finally obtain the image. The two branches utilize the same network architecture U-Net, initialized with the pre-trained weights of Stable Diffusion~\cite{rombach2022ldm}. In detail, our framework mainly consists of three crucial components: 1) Semantic-Aware Appearance Encoder, encoding the multi-scale features of various human parts from reference images; 2) Shared Self-Attention, which obtains detailed information and spatial information by sharing keys and values in self-attention layers between denoising U-Net and appearance encoder, and supports pose control by utilizing a lightweight pose encoder; 3) Enhanced Mask-Guided Subject Selection, achieving precisely subject selection by explicitly introducing subject masks into the self-attention mechanism.

\begin{figure}[!t]
\centering
\includegraphics[width=\linewidth]{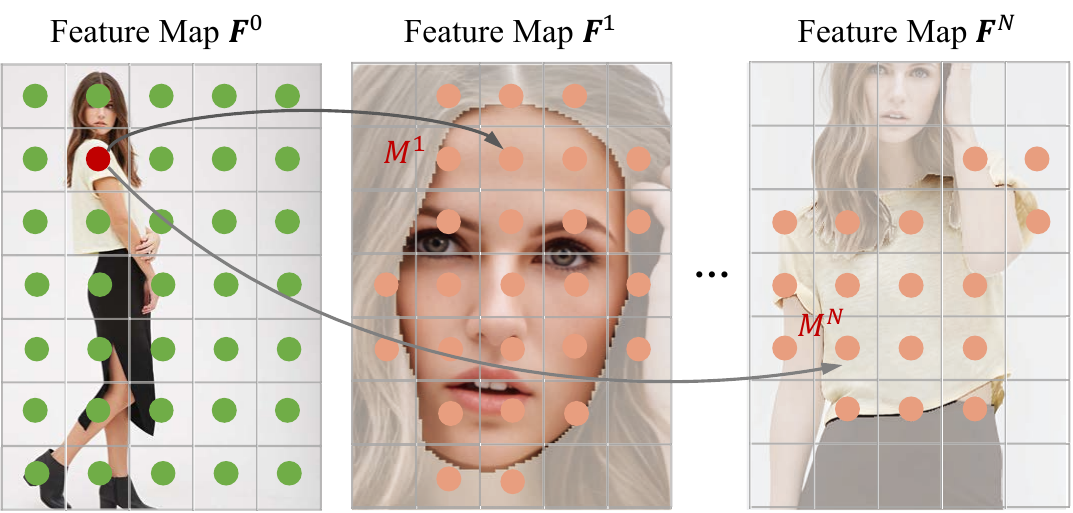}
\caption{\textbf{Illustration of our Mask-Guided Attention.} For each patch $s$ (red point) on the feature map $\bm{F}^{0}$, given subject masks $M^{1:N}$ on the $N$ reference images, we only attend patch $s$ to features in these masks along with the patches on itself.}
\label{fig:mask_guided}
\end{figure}

\cparagraph{Semantic-Aware Appearance Encoder.} In image conditioned generation tasks, previous work~\cite{ye2023ipadapter,yang2023paintbyexample,zhang2024ssrencoder} employs CLIP image encoder~\cite{radford2021clip}, or combined with some simple linear layers to encode reference images, thereby replacing the original text encoder in Stable Diffusion~\cite{rombach2022ldm}. However, such methods struggle to preserve appearance details due to the loss of spatial representations when encoding the reference images into semantic-level features.

Inspired by recent works~\cite{referenceonlycontrol,cao2023masactrl,xu2023magicanimate,hu2023animateanyone} on dense reference image conditioning, we propose a semantic-aware appearance encoder with improved identity and details preservation. Specifically, we adopt a framework identical to the denoising U-Net for the appearance encoder. Unlike the denoising branch, we do not add any noise to the reference images. Given $N$ images capturing various parts of human appearance, we first compress them into latent features and then input them into the copied trainable U-Net. Instead of simply piecing multiple reference images together, we pass the latent features of different parts through the appearance encoder one by one and provide a textual class label for each part. These text labels, such as face, hair, upper body clothes, etc., are converted into feature representations by CLIP text encoder~\cite{radford2021clip} and then injected into the appearance encoder through cross-attention. This simple yet effective external condition provides a classifier-like guidance, which enables the encoder to have semantic awareness of different parts of the human appearance rather than simply performing operations such as image downsampling and upsampling. This helps produce results that are not only rich in detail, but also flexible and realistic.

In the encoding process, we set the timestep to 0 and only perform one processing instead of iterating successively, so it will not cause time burden at the inference stage. We cache the features before each self-attention layer for the next multi-image conditioned generation.

\cparagraph{Shared Self-Attention.} After obtaining the multi-layer feature maps of $N$ reference images, we do not directly add them to the features in denoising U-Net, but use shared keys and values in self-attention to achieve feature injection. This is because our reference and target images are not structurally aligned.

Take one certain self-attention layer as an example. Given the features of $N$ reference images $\bm{F}^{1:N}$ and the feature maps $\bm{F}^{0}$ in the denoising U-Net, we concatenate the feature maps of them side-by-side as input to the self-attention layer, denoted as $[\bm{F}^{0}|\bm{F}^{1}|\cdots|\bm{F}^{N}]$. This allows each location $s$ on $\bm{F}^{0}$ to attend to all locations on itself and reference feature maps, calculated as:
\begin{equation}
\begin{split}
    \tilde{\bm{F}}_{s}^{0} =&\ \text{SoftMax}\left(
    \frac{
    Q(\bm{F}_s^0)\cdot K([\bm{F}^{0}|\bm{F}^{1}|\cdots|\bm{F}^{N}])^{\top}
    }{
    \sqrt{d}
    }\right)  \\
    & \cdot V([\bm{F}^{0}|\bm{F}^{1}|\cdots|\bm{F}^{N}])
    \label{eq:vanilla_self_attn}
\end{split}
\end{equation}

We retain the cross-attention layers in Stable Diffusion~\cite{rombach2022ldm} for injecting CLIP features of reference images and optional text input. We use the decoupled cross-attention proposed by IP-Adapter~\cite{ye2023ipadapter} to support both images and text input. Specifically, feature maps $\tilde{\bm{F}}^{0}$ obtained from shared self-attention serve as the origin of the query, and the reference image features and text features are each used as the key and value of the two cross-attention. The final feature maps are the sum of the two cross-attention outputs.

To further enhance the controllability of the human image generation, we add the pose map as an additional control. We construct a tiny convolution network, which is similar to the condition embedding network in ControlNet~\cite{zhang2023controlnet}, to extract the features of the pose map. The features are then added to the initial feature maps in the denoising U-Net.

\cparagraph{Enhanced Mask-Guided Subject Selection.}
We find that the vanilla shared self-attention leads to interference from irrelevant subjects in the reference images (shown in the 6th column in Fig.~\ref{fig:ablation}), resulting in an unnatural appearance and background. To synthesize human images conditioned on specified parts from each reference image, we enhance the vanilla self-attention mechanism by incorporating subject masks in the reference images. Fig.~\ref{fig:mask_guided} presents this mechanism. Starting with a patch $s$ on a feature map $\bm{F}^{0}$ in the denoising U-Net, and subject masks $M^{1:N}$ on the $N$ reference images. When computing the attention map between the one in the denoising U-Net and those from the appearance encoder, patches that do not lie in these masks are ignored. Hence, the target patch $s$ only has access to features of the subjects specified by masks in the reference images, thereby avoiding interference from other elements such as the background. The final formulation of the mask-guided attention is defined as follows:
\begin{equation}
\begin{split}
    \tilde{\bm{F}}_{s}^{0} =&\ \text{SoftMax}\left(
    \frac{
    Q(\bm{F}_s^0)\cdot K([\bm{F}^{0}|\cdots|\bm{F}^{N}_{M^{N}}])^{\top}
    }{
    \sqrt{d}
    }\right)\\
    & \cdot V([\bm{F}^{0}|\cdots|\bm{F}^{N}_{M^{N}}])
    \label{eq:mask_guided_self_attn}
\end{split}
\end{equation}

In practice, to avoid misalignment between masks and original images caused by downsampling, a full-ones convolutional kernel is applied to the mask before each attention layer, ensuring that the mask preserves critical regions. Overall, the mask-guided attention enhances the ability of Parts2Whole to precisely extract the appearance of specified subjects in reference images.

\section{Experiments}

\begin{figure*}
    \centering
    \includegraphics[width=\textwidth]{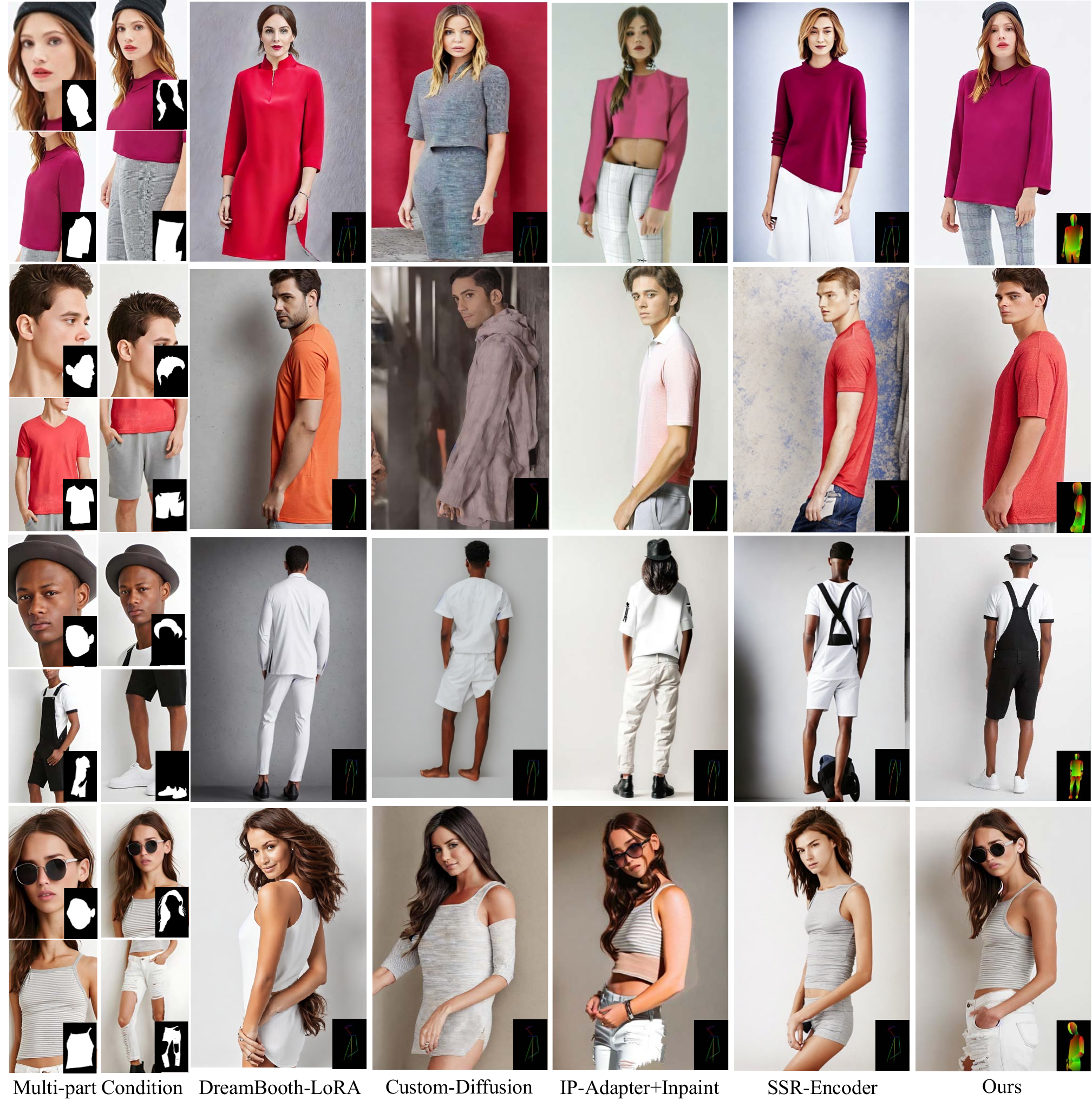}
    \caption{Qualitative results generated by Parts2Whole and existing alternatives on our partitioned test set. We do not show the text condition in the figure, but notably, when we input the reference images to our proposed appearance encoder, we will pass in short labels such as face, hair or headwear, upper body clothes, lower body clothes, whole body clothes, shoes, etc.}
    \label{fig:comparison}
\end{figure*}

\subsection{Implementation Details}

\cparagraph{Dataset.}
To train the Parts2Whole model, we build a multi-modal dataset comprising about 41,500 reference-target pairs from the open-source DeepFashion-MultiModal dataset~\cite{jiang2022text2human,liuLQWTcvpr16DeepFashion}. 
Each pair in this newly constructed dataset includes multiple reference images, which encompass human pose images (e.g., OpenPose, Human Parsing, DensePose), various aspects of human appearance (e.g., hair, face, clothes, shoes) with their short textual labels, and a target image featuring the same individual (ID) in the same outfit but in a different pose, along with textual captions.

The DeepFashion-MultiModal dataset exhibits noise in its ID data. For example, different images are tagged with the same ID but depict different individuals. To address this issue, we first cleanse the IDs by extracting facial ID features from images tagged with the same ID using InsightFace\cite{deng2018arcface, InsightFaceProject}. Cosine similarity is then used to evaluate the similarity between image ID feature pairs to distinguish between different ID images within the same ID group. Subsequently, we utilize DWPose\cite{yang2023dwpose} to generate pose images corresponding to each image. Guided by human parsing files, we crop human images into various parts. Due to the low resolution of the cropped parts, we apply Real-ESRGAN\cite{wang2021realesrgan} to enhance the image resolution, thus obtaining clearer reference images. Textual descriptions of the original dataset are used as captions. For constructing pairs, we select images with cleaned IDs that feature the same clothes and individual but in different poses. Specifically, a pair contains multiple parts from one human image as reference images, and an image of the person in another pose as the target. Finally, we build a total of about 41,500 pairs, of which the training set is about 40,000 and the test set is about 1,500 pairs.

\cparagraph{Detailed Configurations.}
In this work, the denoising U-Net and the appearance encoder both leverage the pre-trained weights from Stable Diffusion-1.5~\cite{rombach2022ldm}. We use CLIP Vision Model with projection layers as our image encoder, initialized with Stable Diffusion Image Variations~\cite{sdimagevariation}. During training, we set the initial learning rate 1e-5 with a batch size of 64. The model is trained using 8 A800 GPUs, for a total of 30000 iterations. To maintain the capability of image generation, we randomly drop all of the reference image features and the pose condition with a probability of 0.2. At the same time, to improve the flexibility of generation, we randomly drop each appearance condition with a probability of 0.2, so that the human images can be generating from indefinite reference images. At the inference stage, we adopt DDIM sampler~\cite{song2020ddim} with 50 steps, and set the guidance scale to 7.5.

\subsection{Comparison with Existing Alternatives}

Our Parts2Whole targets at controllable human image generation conditioned on multiple parts of human appearance. To evaluate the performance of our proposed framework, we compare our Parts2Whole with existing subject-driven solutions. For fairness, we make some improvements to the methods, to make them more suitable for generating human images from multiple conditions.

\cparagraph{Test-time Fine-tuning Methods.}
Among the tuning-based methods, we adopt DreamBooth LoRA~\cite{ruiz2023dreambooth,hu2021lora} and Custom Diffusion~\cite{kumari2023customdiffusion} as baseline methods for comparison, as these methods are relatively robust and effective. DreamBooth LoRA inserts a smaller number of new weights into Stable Diffusion~\cite{rombach2022ldm} and only trains these parameters on just a few images of a subject or style, thereby associating a special word in the prompt with the example images. Custom Diffusion fine-tunes only key and value projection matrices in the cross-attention layers to customize text-to-image models. Given as input several aspects of human appearance, we use these two methods to fine-tune Stable Diffusion, such that it learns to bind identifiers with specific human parts. As shown in \cref{fig:comparison}, when it comes to multi-aspect composition, the attributes of different parts in images generated by these tuning-base methods mix together, resulting in unrealistic human images. In contrast, Parts2Whole generates high-fidelity results without the need for parameter tuning.

\cparagraph{Reference-based Methods.}
Among the tuning-free methods, we adopt IP-Adapter~\cite{ye2023ipadapter} and SSR-Encoder~\cite{zhang2024ssrencoder} for comparison. IP-Adapter is an image prompt adapter that can be plugged into diffusion models to enable image prompting, and can be combined with other adapters like ControlNet~\cite{zhang2023controlnet}. We firstly use IP-Adapter FaceID and ControlNet to generate human images from facial appearance and pose maps. Then we repaint hair, clothes, shoes and other areas using the specific image step by step, thereby achieving multi-image conditioned generation of portraits in a multi-step way. SSR-Encoder is an effective encoder designed for selectively capturing any subject from single or multiple reference images by the text query or mask query. For fairness, we fine-tune it in our human dataset to enhance its ability for human images.

We compare our Parts2Whole with the above two reference-based alternatives in the test set. For quantitative comparison, we compute the commonly used CLIP score and DINO score to evaluate the similarity between the generated image and the specified human parts. For further alignment evaluation, we use DreamSim~\cite{fu2023dreamsim}, a new metric for perceptual image similarity that bridges the gap between ``low-level'' metrics (e.g., LPIPS, PSNR, SSIM) and ``high-level'' measures (e.g., CLIP). Since the generated image is conditioned on multiple parts of human appearance, it is difficult to evaluate the degree of alignment by calculating the average metric with these multiple images. Therefore, we calculate the above three indicators between the output image and the original reference portrait from which these different parts come. We present the quantitative results in \cref{tab:quan_comp} and the qualitative results in ~\cref{fig:comparison}. Both IP-Adapter and SSR-Encoder fail to maintain alignment with the specified appearance images and often produce unrealistic results when multi-part combinations are involved. In comparison, our method achieves the best results in terms of image quality and appearance alignment.

\begin{table}\small
  \centering
  \caption{Quantitative comparison between our Parts2Whole and existing reference-based alternatives.}
  \begin{tabular}{@{}lccc@{}}
    \toprule
    Method & CLIP$\uparrow$ & DINO$\uparrow$ & DreamSim\cite{fu2023dreamsim}$\downarrow$ \\
    \midrule
    IP-Adapter~\cite{ye2023ipadapter} + Inpaint & 80.1 & 69.8 & 0.445   \\
    SSR-Encoder~\cite{zhang2024ssrencoder} & 86.9 & 75.1 & 0.346 \\
   Parts2Whole (Ours) & \textbf{91.2} & \textbf{93.7} & \textbf{0.221}  \\
    \bottomrule
  \end{tabular}
  \label{tab:quan_comp}
\end{table}

\cparagraph{User Study.}
We conduct a user study to further evaluate the reference-based methods IP-Adapter~\cite{ye2023ipadapter}, SSR-Encoder~\cite{zhang2024ssrencoder} and our Parts2Whole. We randomly select 20 pairs of reference-target pairs from the test set. For each pair, we provide multiple referential appearance images, pose images, textual captions, and the generated human images. We evaluate the performance from two main aspects: first, the \textbf{quality} of the generated images, which primarily refers to the realism, rationality, and clarity of the images; and second, the \textbf{similarity} between the generated images and the reference images. The similarity assessment includes consistency in ID, pose, texture, and color between the generated images and the reference images. We involve 20 users in the user study, who are required to score the three methods based on these two evaluative aspects. The final experimental results are shown in \cref{tab:user_study}, from which we observe that our model owns obvious superiorities for alignment with given appearance conditions.

\begin{table}
  \centering
  \caption{User study on the comparison with existing reference-based alternatives. ``Quality'' and ``Similarity'' measures synthesis quality and appearance preservation. Each metric is rated from 1 (worst) to 5 (best).}
  \begin{tabular}{@{}lcccc@{}}
    \toprule
    Method & Quality$\uparrow$ & Similarity$\uparrow$ \\
    \midrule
    IP-Adapter~\cite{ye2023ipadapter} + Inpaint &  3.78 & 3.58  \\
    SSR-Encoder~\cite{zhang2024ssrencoder} & 3.64  & 3.14 \\
   Parts2Whole (Ours) & \textbf{4.52} & \textbf{4.55} \\
    \bottomrule
  \end{tabular}
  \label{tab:user_study}
\end{table}

\subsection{Ablation Studies}

\begin{figure*}
    \centering
    \includegraphics[width=\textwidth]{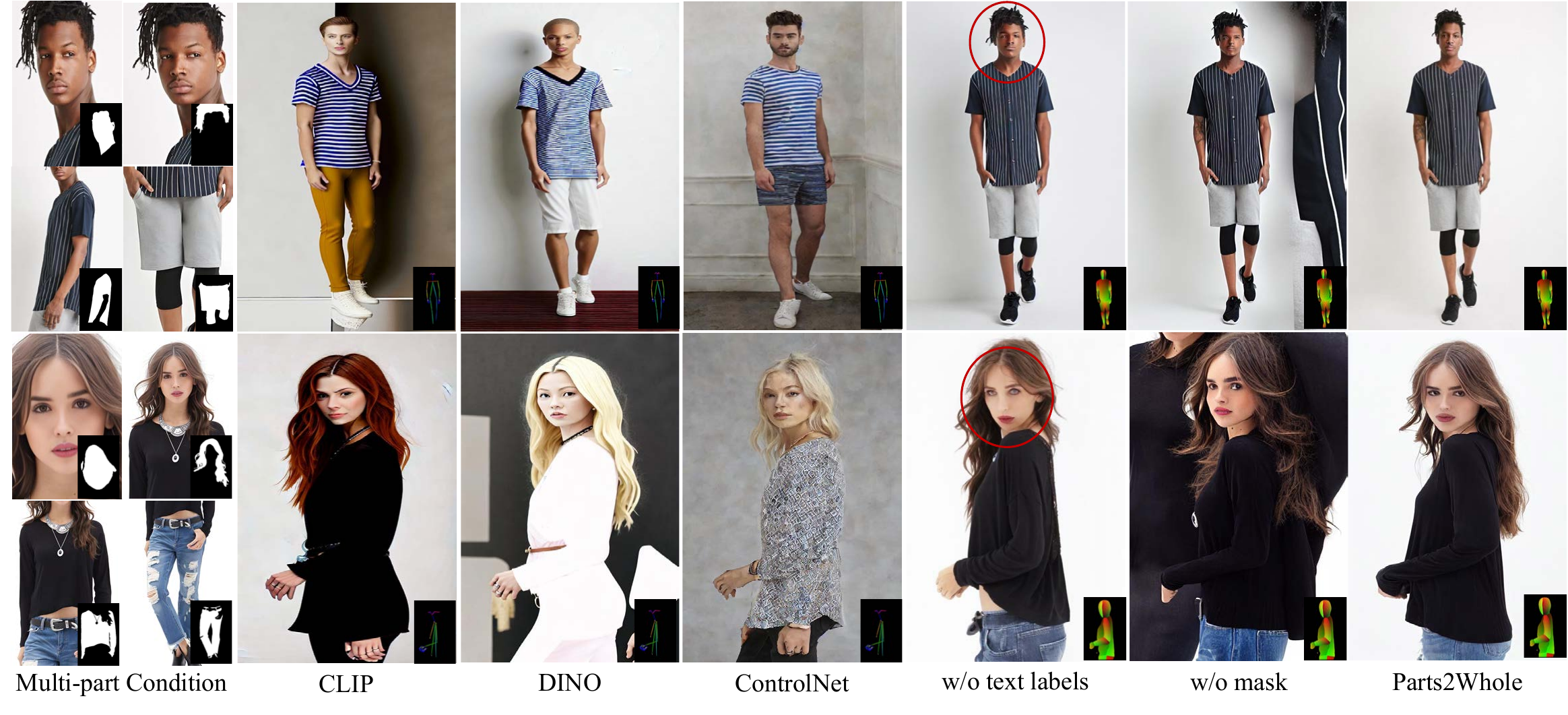}
    \caption{Qualitative analysis of using different backbones for the appearance encoder, and our proposed methods.}
    \label{fig:ablation}
\end{figure*}

\cparagraph{Appearance Encoder for Multiple Images.}
As described in Sec.~\ref{subsec:unifiedref}, to extract detailed features from multiple reference images, our Parts2Whole designs an appearance encoder by copying the network structure and pre-trained weights from the denoising U-Net. Here, we compare it to the baseline with other image encoders. Specifically, we leverage the CLIP image encoder~\cite{radford2021clip}, DINOv2~\cite{oquab2024dinov2}, and ControlNet~\cite{zhang2023controlnet} as feature extractors and apply the same training settings for fair comparison. The qualitative results of generated human images are presented in \cref{fig:ablation}. From the second and third columns of the figure, we observe that these semantic-level feature extractors cannot preserve the appearance details of multiple reference images and only extract color and rough texture. ControlNet directly adds different image features with misaligned structures to the feature maps, resulting in unstable image quality. In contrast, our proposed appearance encoder provides fine-grained details of multiple aspects of human appearance.

\cparagraph{Semantic-Aware Encoder.} In the process of encoding multiple reference image features, we provide a textual class label for each aspect of human appearance, thus providing a classifier-like guidance. To assess the effectiveness of the additional external condition, we compare it with directly concatenating multiple reference images in the dimension of width as input to the appearance encoder. As shown in the 5th column in \cref{fig:ablation}, simply piecing reference images produces images relatively aligned with the given images, but leads to stiff-looking and unrealistic results. This is because modeling only the image itself makes the model lack awareness of different types of appearances. Conversely, after injecting different semantic labels for each reference image, the model has an awareness of various parts of the human appearance, producing realistic and flexible portraits.

\cparagraph{Mask-Guided Subject Selection.}
To precisely select subjects from multiple reference images, we introduce the subject masks into the shared self-attention mechanism. To evaluate the effectiveness of our proposed mask-guided attention, we compare it with that without masks. When not using subject masks, the image is generated with reference to all patches of the conditional images, including the unexpected background or other parts. As shown in \cref{fig:ablation}, due to the generation being interfered with by irrelevant subjects, the model produces homogeneous colors or appears with unexpected backgrounds or subjects. On the contrary, with the support of mask-guided attention, Parts2Whole accurately refers to the appearance of the specified parts to generate real human images.

\begin{table}\small
  \centering
  \caption{Quantitative analysis of using semantic-aware encoder and mask-guided subject selection.}
  \begin{tabular}{@{}lcccc@{}}
    \toprule
    Method & CLIP$\uparrow$ & DINO$\uparrow$ & DreamSim\cite{fu2023dreamsim}$\downarrow$ & FID$\downarrow$ \\
    \midrule
    w/o text labels & 90.1 & 91.9 & 0.248 & 23.95 \\
    w/o mask & 90.8 & 91.6 & 0.243 & 19.79 \\
   Parts2Whole & \textbf{91.2} & \textbf{93.7} & \textbf{0.221} & \textbf{17.29} \\
    \bottomrule
  \end{tabular}
  \label{tab:abla}
\end{table}

\begin{figure*}
    \centering
    \includegraphics[width=\textwidth]{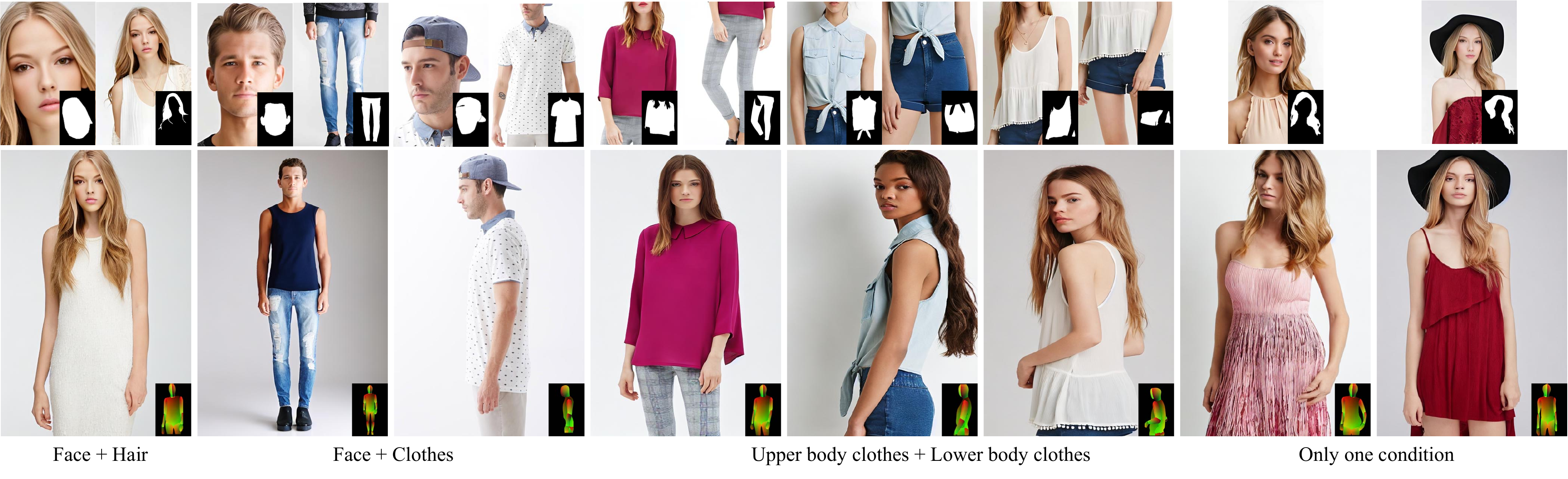}
    \caption{The generated results from combinations of a different number of conditions.}
    \label{fig:any_quantity}
\end{figure*}

\subsection{More Results}
\begin{figure*}
    \centering
    \includegraphics[width=\textwidth]{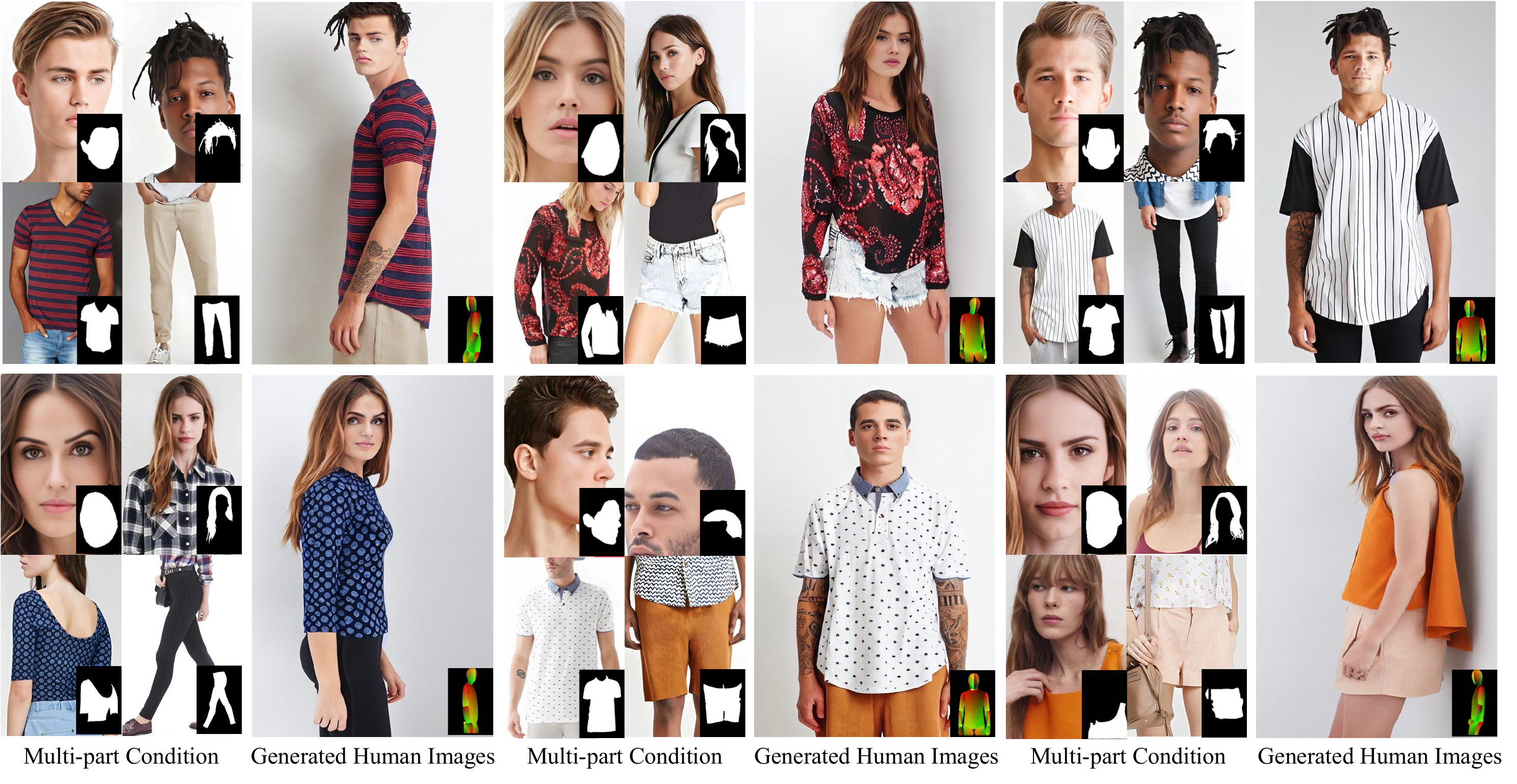}
    \caption{Results of image generation using selected parts from different individuals as control conditions.}
    \label{fig:any_person}
\end{figure*}

\cparagraph{Body Parts of Any Quantity.}
Our Parts2Whole is able to generate human images from varying numbers of condition images, such as single hair or face input, or arbitrary combinations like ``Face + Hair'', ``Face + Clothes'', and ``Upper body clothes + Lower body clothes''. The experimental results are presented in \cref{fig:any_quantity}. The generated results under different control condition combinations still maintain high quality and realism. This flexibility enables our method to have broader application.

\cparagraph{Multiple Parts from Different Humans.}
We select various parts from different human images to serve as conditional images. For example, the face from person A, the hair or headwear from person B, the upper clothes from person C, and the lower clothes from person D. These parts are collectively used as control conditions for generation. The experimental results, as shown in \cref{fig:any_person}, demonstrate that our method not only accurately maps different parts of the reference image to the corresponding regions in the target image but also effectively preserves the details of the conditions, producing realistic images.

\section{Conclusion}

In this work, we propose Parts2Whole, a novel framework for controllable human image generation conditioned on multiple reference images, including various aspects of human appearance (e.g., hair, face, clothes, shoes, etc.) and pose maps. Based on a dual U-Net design, we develop a semantic-aware appearance encoder to process each condition image with its label into multi-scale feature maps and inject those detail-rich reference features into the generation via a shared self-attention mechanism. This design retains details from multiple references and looks very good. We also enhance vanilla self-attention by incorporating subject masks, enabling Parts2Whole to synthesize human images from specified parts from condition images. Extensive experiments demonstrate that our Parts2Whole performs well in terms of image quality and condition alignment.

\cparagraph{Future Works.} Our Parts2Whole is currently trained at the resolution of 512, which may cause artifacts in some generated results. This could be improved by using higher resolutions and larger diffusion models like SD-XL~\cite{podell2023sdxl} as our backbone. Furthermore, it will be valuable to achieve the try-on of layer-wise clothing based on our Parts2Whole.

{
    \small
    \bibliographystyle{ieeenat_fullname}
    \bibliography{arxiv}
}

\appendix
\clearpage
\setcounter{page}{1}
\maketitlesupplementary

\section{Dataset}

\begin{figure*}[t]
    \centering
    \includegraphics[width=0.95\textwidth]{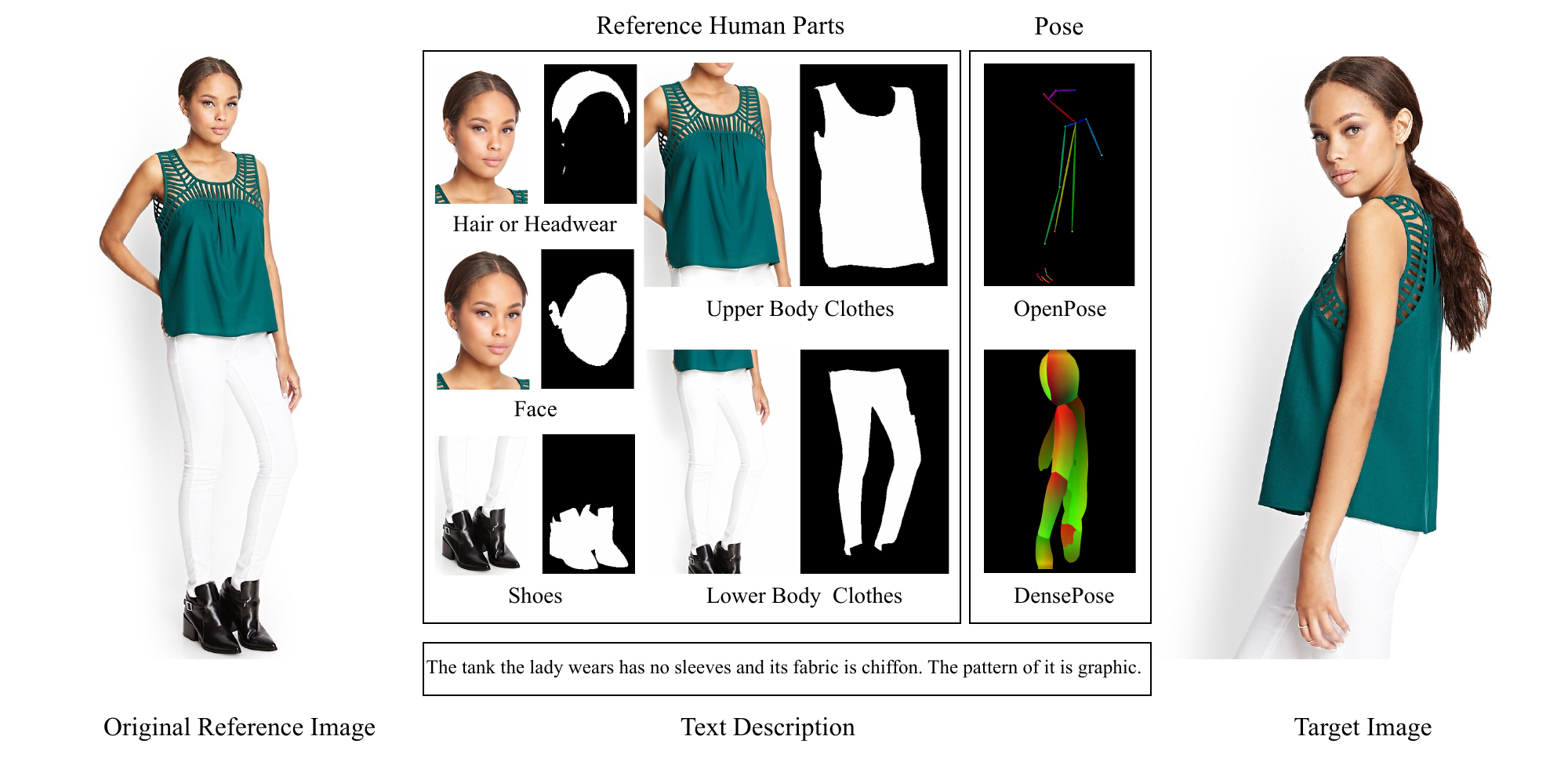}
    \caption{A sample of reference-target image pair in our dataset. Reference Human Part images are obtained from the reference image based on the human parsing image, while the pose and text description are descriptions of the target image.}
    \label{fig:sample_data}
\end{figure*}

\begin{figure*}[ht!]
    \centering
    \includegraphics[width=\textwidth]{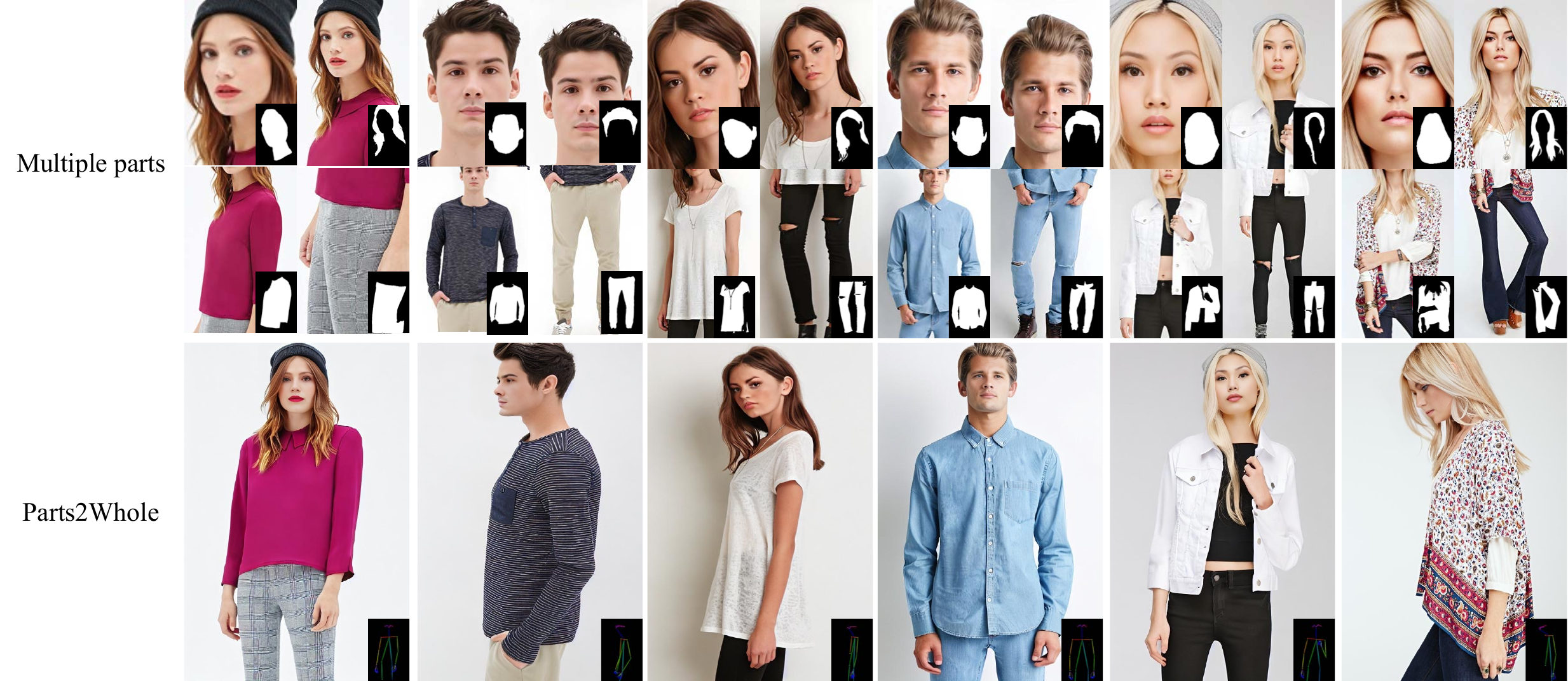}
    \caption{Generated human images with OpenPose as pose condition by Parts2Whole. The upper row displays multiple reference human parts, while the lower row shows the results generated by our Parts2Whole method under the control of reference images and OpenPose image.}
    \label{fig:supp_openpose}
\end{figure*}

Generating the human body from multiple conditional parts is a significant undertaking, but lacking a directly available dataset. The datasets related to this task, such as those for Virtual Try-on\cite{choi2021viton,morelli2022dress,zablotskaia2019dwnet}, primarily suffer from a lack of multiple controllable conditions and are often limited to single control conditions (clothing). Issues with these datasets include a limited variety of clothing types, absence of facial data, low resolution, and lack of textual captions. The DeepFashion-MultiModal dataset~\cite{jiang2022text2human,liuLQWTcvpr16DeepFashion} aligns more closely with our task as it includes a vast array of human body images, the same person and same clothes in different poses, and precise human parsing labels. However, this dataset cannot be used directly and requires data cleansing and further post-processing.

\cparagraph{ID Cleansing.}
In the DeepFashion-MultiModal dataset, there is some confusion with IDs where images of different individuals are mistakenly labeled under the same ID. We start by cleansing these IDs, extracting facial ID features from images tagged with the same ID using InsightFace\cite{deng2018arcface, InsightFaceProject}. Cosine similarity is then used to evaluate the similarity between image ID feature pairs, allowing us to reclassify IDs within the same ID group. After cleansing, images from the same ID and the same clothes are selected if there are two or more images available.

\cparagraph{Building Reference-Target Pair.}
We use images with human parsing labels from the dataset as reference images. Target images are then selected from the same ID and clothing, creating pairs with the reference image.

\cparagraph{Obtaining Reference Human Part.}
We crop the images according to the provided human parsing labels. Specifically, we divide the human image into six parts: upper body clothes, lower body clothes, whole body clothes, hair or headwear, face, and footwear. Each part is cropped according to the human parsing labels to obtain the crop image and corresponding mask image. Due to the low resolution of the cropped parts, we apply Real-ESRGAN\cite{wang2021realesrgan} to enhance the image resolution, thus obtaining clearer reference images.

\cparagraph{Obtaining Target Description.}
Based on the reference human parts, we need to generate images that resemble the target image, requiring a description of the target image. The description is divided into two parts: one for the human body's pose and another for the target image's textual description. For pose information, we utilize DWPose\cite{yang2023dwpose} to generate pose images corresponding to each image, and for DensePose, we use the provided DensePose files from the dataset. The textual description for each image is taken directly from the dataset's accompanying text description.

\cparagraph{Introduction to the Final Dataset.}
Finally, we have constructed a multimodal dataset with approximately 41,500 reference-target pairs derived from the open-source DeepFashion-MultiModal dataset~\cite{jiang2022text2human,liuLQWTcvpr16DeepFashion}. The controllable conditions for each pair are categorized into two main types. The first type is the appearance reference image, which is subdivided into six parts: upper body clothes, lower body clothes, whole body clothes, hair or headwear, face, and footwear. Each image is accompanied by a corresponding mask and has undergone super-resolution processing. These data elements are sourced from the original reference image. The second type is the target description, primarily consisting of pose and text description. The pose is further divided into OpenPose and DensePose, all of which are derived from the target image. A sample of reference-target image pair in our dataset is shown in Fig.~\ref{fig:sample_data}.

\section{Different Types of Pose Maps}

\begin{figure*}[h]
    \centering
    \includegraphics[width=1\textwidth]{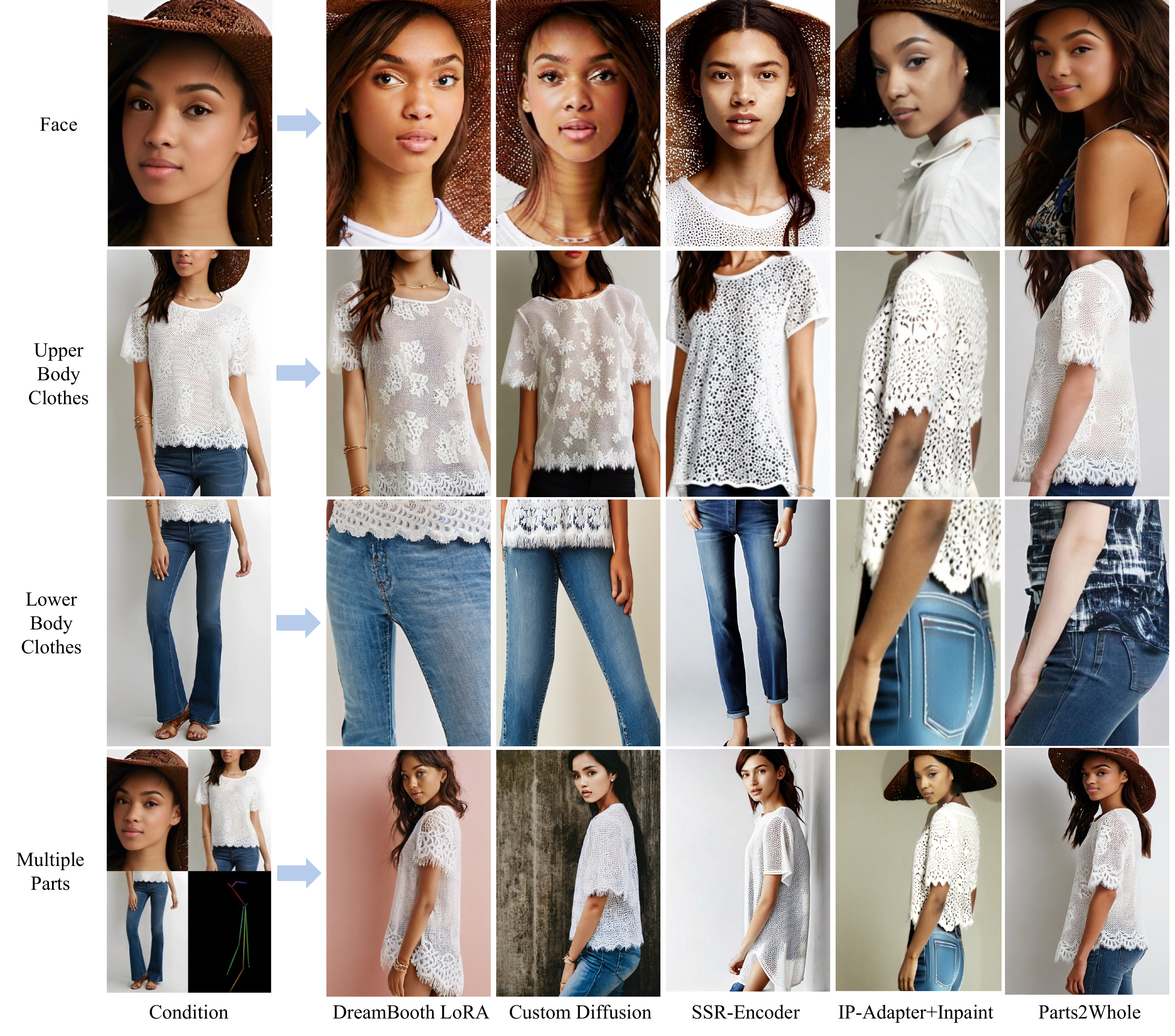}
    \caption{Results of single-condition and multi-condition generation. The top three rows labeled ``face'', ``upper body clothes'', ``lower body clothes'' represent separate conditions. The fourth row demonstrates joint control under multi-part conditions.}
    \label{fig:existing_methods}
\end{figure*}

To show the ability of Parts2Whole to generate human image conditions on different types of pose maps, we train a new Parts2Whole model but with \textbf{OpenPose} as a condition. As shown in \cref{fig:supp_openpose}, the generated images strictly maintain consistency with the target pose, and each body part retains the appearance information from the reference images.

\section{Discussion about Existing Methods}

In our main text, we conduct experiments with both tuning-based methods DreamBooth LoRA~\cite{ruiz2023dreambooth,hu2021lora} and Custom Diffusion~\cite{kumari2023customdiffusion} and tuning-free methods such as IP-Adapter~\cite{ye2023ipadapter} and SSR-Encoder~\cite{zhang2024ssrencoder}. The results of these experiments are further illustrated in Fig~\ref{fig:existing_methods}, which showcases the performance of these methods conditioned on both a single image and multiple images.

The results demonstrate that while these methods generally perform well under a single control condition, they exhibit significant issues when multiple conditions are applied. For example, DreamBooth LoRA encountered cases where pants were omitted, and both Custom Diffusion and SSR-Encoder show alterations in facial ID. The phenomenon observed is attributed to the spatial misalignment of the input multi-body parts with the target image, and the lack of specific design in existing methods to address the variation in spatial positions during feature injection. For instance, methods like SSR-Encoder, Custom Diffusion, and IP-Adapter incorporate features into the denoising UNet through cross-attention mechanisms. They encode reference images into other modal features (e.g. semantic features) and utilize the cross-attention keys ($K$) and values ($V$) from them rather than from \textbf{image dimensional} feature maps. In this process, the correlation between the reference images and the target image \textbf{loses the spatial relationship} of the original image dimensions. It is difficult for these methods to effectively model the attention from various conditional feature maps at different locations in the target image, resulting in a mixture of attributes from different subjects.

Conversely, our Parts2Whole model employs shared self-attention between the reference features and the feature maps in the Denoising U-Net, executed on the image dimension. This allows our model to establish a more precise correlation between different condition images and distinct positions within the feature maps, thereby generating results that are consistent with the detailed attributes of multi-condition images.

\end{document}